\documentclass{article} 
\usepackage{iclr2023_conference,times}

\usepackage{amsmath,amsfonts,bm}

\def\eqref#1{equation~\ref{#1}}

\def\1{\bm{1}}

\def\vx{{\bm{x}}}

\def\vz{{\bm{z}}}

\def\mE{{\bm{E}}}

\DeclareMathAlphabet{\mathsfit}{\encodingdefault}{\sfdefault}{m}{sl}
\SetMathAlphabet{\mathsfit}{bold}{\encodingdefault}{\sfdefault}{bx}{n}

\newcommand{\R}{\mathbb{R}}

\usepackage{hyperref}
\usepackage{url}
\usepackage{wrapfig}
\usepackage{xspace}
\usepackage{enumerate}
\usepackage{enumitem}
\usepackage{graphicx}
\usepackage{makecell}
\usepackage{algorithm}
\usepackage[frozencache,cachedir=.]{minted} 
\usepackage{lipsum}
\usepackage{colortbl}
\definecolor{mygray}{gray}{.93}
\usepackage{transparent}

\makeatletter
\DeclareRobustCommand\onedot{\futurelet\@let@token\@onedot}
\def\@onedot{\ifx\@let@token.\else.\null\fi\xspace}

\def\eg{\emph{e.g}\onedot} 
\def\ie{\emph{i.e}\onedot}

\newcommand\figcaption{\def\@captype{figure}\caption} 
\newcommand\tabcaption{\def\@captype{table}\caption}

\makeatother

\title{AIM: Adapting Image Models for Efficient Video Action Recognition}

\author{Taojiannan Yang$^{1}$\thanks{Work done during an internship at Amazon Web Services.}, Yi Zhu$^2$, Yusheng Xie$^2$, Aston Zhang$^2$, Chen Chen$^1$, Mu Li$^2$ \\
	$^1$University of Central Florida \quad $^2$Amazon Web Services
}

\iclrfinalcopy 
\begin{document}
	
	\maketitle
	
	\begin{abstract}
		Recent vision transformer based video models mostly follow the ``\textit{image pre-training then finetuning}" paradigm and have achieved great success on multiple video benchmarks. 
		However, full finetuning such a video model could be computationally expensive and unnecessary, given the pre-trained image transformer models have demonstrated exceptional transferability. 
		In this work, we propose a novel method to Adapt pre-trained Image Models (AIM) for efficient video understanding.
		By freezing the pre-trained image model and adding a few lightweight Adapters, we introduce spatial adaptation, temporal adaptation and joint adaptation to gradually equip an image model with spatiotemporal reasoning capability.
		We show that our proposed AIM can achieve competitive or
		even better performance than prior arts with substantially fewer tunable parameters on four video action recognition benchmarks.
		Thanks to its simplicity, our method is also generally applicable to different image pre-trained models, which has the potential to leverage more powerful image foundation models in the future. The project webpage is \url{https://adapt-image-models.github.io/}.
	\end{abstract}
	
	\section{Introduction}
	The ``pre-training then finetuning'' paradigm has played an important role in computer vision. 
	The key to this paradigm is a well pre-trained image model, which can provide strong transferability to downstream tasks through finetuning. 
	Recently, large foundation models \citep{clip, yuan2021florence, tong2022videomae,align, beitv3} can even demonstrate remarkable few-/zero-shot performance  given their learned superior visual representations.
	
	In video understanding, a common practice is also bootstrapping from an image pre-trained model and then finetuning on the video data. 
	There are two dominating directions as shown in Fig.\ \ref{fig:teaser}, one is to extend an image model with additional temporal module \citep{lin2019tsm,zhu2019hidden,arnab2021vivit}, the other is to inflate an image model to a video model \citep{i3d, liu2022videoswin}.
	However, there exists at least two drawbacks for the aforementioned methods.
	First, most approaches require full finetuning (\ie, updating all the model parameters during training) to achieve promising results on common video benchmarks. 
	This is quite costly in terms of both computation and memory footprint, \eg, 1200 Tesla V100 GPU hours to train \citet{liu2022videoswin}.
	Second, it also remains questionable that whether it is necessary to full finetune pre-trained image models given that they have demonstrated excellent transferability. An inadequate finetuning on downstream data might destroy the well generalized representations from such foundation models. 
	
	To overcome the drawbacks, a research direction termed parameter-efficient transfer learning has been trending in natural language processing (NLP) \citep{adapter, prompttuning, ben-zaken-etal-2022-bitfit,hu2022lora}. 
	The goal is to only finetune a small number of (extra) parameters while keeping large pre-trained language models \citep{devlin2018bert, gpt} frozen to attain strong performance. 
	With the rise of large vision transformer (ViT) models, such techniques have been recently introduced to computer vision for efficient transfer learning. 
	However, existing works either focus on tuning a pre-trained image model for image tasks (image-to-image) \citep{visualpixelprompt, convadapter, vpt}, or tuning a pre-trained video model for video tasks (video-to-video) \citet{chen2022adaptformer}. 
	Directly leveraging pre-trained image models for efficient transfer learning to video tasks (image-to-video) is less explored, because image models lack the capability of temporal reasoning.

	In this work, we introduce a new way to \textbf{A}dapt pre-trained \textbf{I}mage transformer \textbf{M}odels (AIM) for efficient video action recognition.
	By freezing the pre-trained image model and adding a few lightweight adapters \citep{adapter} during finetuning, we show that our proposed AIM can achieve competitive or even better results than previous state-of-the-art methods with substantially fewer tunable parameters (Fig.\ \ref{fig:teaser} right). 
	To be specific, we first introduce adapter after self-attention layer in a transformer block to perform \textit{spatial adaptation}. We show that a well pre-trained image model is sufficiently good for spatial modeling in video understanding.
	Then for temporal modeling, we simply reuse the image pre-trained self-attention layer but apply it to the temporal dimension of video input, forcing it to model the relationship across different frames.
	An adapter is also appended for \textit{temporal adaptation}. 
	Finally, we perform \textit{joint adaptation} by adding another adapter in parallel to the MLP layer in a transformer block.
	To summarize, we make the following contributions:

	\begin{figure*}[t]
		\vspace{-5pt}
		\centering
		\includegraphics[width=0.95\linewidth]{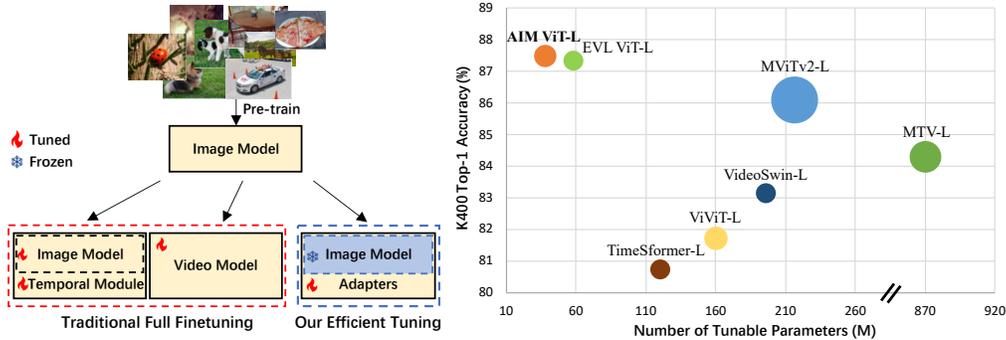}
		\vspace{-6pt}
		\caption{\textbf{Left}: Pipeline comparison between traditional full finetuning  and our efficient finetuning. \textbf{Right}: Performance comparison on K400 dataset \citep{kay2017kinetics}. Bubble size indicates GFLOPS at inference time. Our proposed AIM achieves the highest accuracy while enjoying significantly less number of tunable parameters and GFLOPS.}
		\label{fig:teaser}
	\end{figure*}

	\setlength{\leftmargini}{12pt}
	\begin{enumerate}
		\item We propose a new way to adapt pre-trained image transformer models for efficient video understanding. Our method is generally applicable to different image pre-trained models, simple to implement, and cost-effective to train.
		\item Our method is significantly more efficient than full finetuning a video model, \eg,
		on Swin-B backbone, we can reduce the memory footprint by $50\%$ and training time by $42\%$ compared to VideoSwin \citep{liu2022videoswin}.
		\item AIM achieves comparable or higher performance than previous full finetuned state-of-the-arts on 4 video action recognition benchmarks, \eg, $87.5\%$ on K400 with 38M tunable parameters.
		\item Our method also brings data efficiency,  \eg, AIM outperforms counterpart TimeSformer \citep{timesformer} by $9\%$ absolute accuracy improvement when using  $1\%$ of the training data.
	\end{enumerate}
	
	\section{Related Work}
	
	\textbf{Image pre-trained models.} 
	ViT \citep{vit} and its variants \citep{liu2021swin, wang2021pyramidvit, yuan2021tokenstotoken, dong2022cswin} have been proposed to achieve state-of-the-art performance on image recognition. 
	Once trained, these models could also serve as good initialization for transfer learning to downstream tasks.
	In terms of training techniques, they are commonly trained on large-scale labeled datasests \citep{deng2009imagenet,jft300m,zhai_cvpr2022_scalingvit} in a supervised manner.
	To alleviate the labeling cost, self-supervised learning methods \citep{mocov3,beit,zhou2021ibot,mae,xie2022simmim} are introduced to learn effective representations from unlabeled data. 
	Recent works \citep{clip, align, yuan2021florence,beitv3} adopt large-scale multimodal data (\eg, image-text pairs) for model training, which leads to even more powerful visual representations. 
	In this work, thanks to the simplicity of our proposed method, we could take advantage of these well pre-trained image models and adapt them efficiently to solve video tasks.
	
	\textbf{Video action recognition.} 
	A paradigm shift from using convolutional networks \citep{i3d,r21d,yang2021mutualnet,lin2019tsm,feichtenhofer2019slowfast} to transformers has been observed for video action recognition. Most works use image pre-trained models as initialization and extend them to video models by introducing new temporal modules \citep{timesformer, arnab2021vivit, zhang2021vidtr, yan2022multiview} or inflate them to video models \citep{liu2022videoswin}. 
	Another direction is to directly pre-train a video model in a self-supervised manner \citep{kuang2021video, feichtenhofer2022videomae, zolfaghari2021crossclr, tan2021vimpac}.
	However, all these models are full finetuned on video data, which makes the training cost unaffordable to most researchers and practitioners. 
	There are some recent works \cite{xclip, promptclip} extending CLIP to perform action recognition, but they are multimodal methods which requires additional text branch. 
	Our proposed AIM leverages existing pre-trained image models (no need for video model pre-training), only tunes a small number of model parameters (much more efficient than full finetuning), and achieves comparable or even better performance than previous state-of-the-arts.

	\textbf{Parameter-efficient finetuning} techniques \citep{adapter,hu2022lora,prompttuning,li-liang-2021-prefix,unifiedadapter,ben-zaken-etal-2022-bitfit,fixsparsemask,qing2022mar} are first proposed in NLP since full finetuning the increasingly larger language models for various downstream tasks becomes less feasible. 
	Their goal is to reduce the number of trainable parameters thus lowering the computation cost, while reaching or surpassing the performance of full finetuning. 
	Recently, parameter-efficient transfer learning is also studied in computer vision \citep{vpt,visualpixelprompt,chen2022adaptformer,convadapter,gao2022dept}. 
	All these methods focus on adapting models in the same domain (\eg, image-to-image or video-to-video), while our method adapts an image model for video tasks. 
	One concurrent work \citep{frozenclip} also studies how to adapt image pre-trained models for video action recognition. However, there are several major differences. First, they add new trainable decoder branches, which consist of 3D convolutions and cross-frame attention, to the frozen image encoder. We simply reuse image pre-trained self-attention to perform temporal modeling, while enjoying better performance and less tunable parameters. Second, our method is shown to be compatible with different image models, while \citet{frozenclip} only shows its effectiveness on CLIP image encoder.
	
	\section{Methodology}
	\label{sec:method}
	
	In this section, we first briefly describe ViT and video baselines (Sec. \ref{subsec:preliminary}). Then we introduce spatial adaptation (Sec. \ref{subsec:spatial adapter}), temporal adaptation (Sec. \ref{subsec:temporal module}) and joint adaptation (Sec. \ref{subsec:joint module}), to show how we adapt a pre-trained image model for effective video modeling step-by-step.
	
	\subsection{Preliminary}
	\label{subsec:preliminary}
	Since Vision Transformer (ViT) \citep{vit} is proposed, transformer-based models have been widely adopted in various computer vision tasks, including video action recognition. 
	In this work, we focus on adapting pre-trained image transformer models and compare to full finetuned video transformer models, unless otherwise stated. 
	
\begin{figure*}[t]
		
		\centering
		\includegraphics[width=0.85\linewidth]{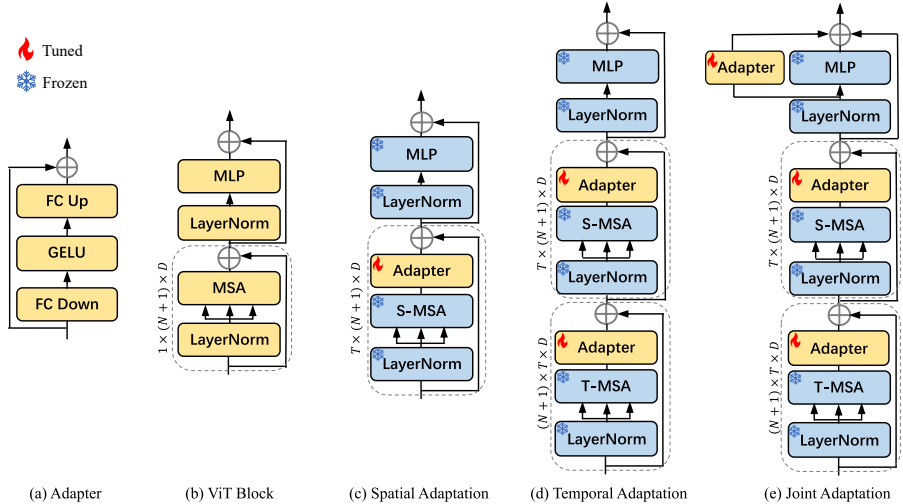}
		\vspace{-2pt}
		\caption{We show how we adapt a standard ViT block (b) for video action recognition, by gradually adding spatial adaptation (c), temporal adaptation (d) and joint adaptation (e). 
			Note that S-MSA and T-MSA share weights but are applied to different input dimensions. During training, only newly added Adapters are updated while all the other layers are frozen.}
		\label{fig:adapter}
		\vspace{-20pt}
	\end{figure*}

	More specifically, ViT handles an image as a sequence of small patches. 
	Given input image $\vx \in \R^{H \times W \times C}$, ViT first splits the image to $N$ non-overlapping patches and maps each patch to a $D$-dim patch embedding via a trainable linear projection \citep{qian2021blending,qian2022makes}. 
	Here, $(H, W)$ is the image resolution and $C$ is the number of channels. Patch embeddings $\vx_p \in \R^{N \times D}$, where $N = HW/P^2$ and $P$ denotes the patch size. 
	Then a learnable [class] token is prepended to $\vx_p$ as $\vx_0 = [\vx_{class};\: \vx_p] \in \R^{(N+1) \times D}$. 
	To encode positional information, positional embeddings $\mE_{pos} \in \R^{(N+1) \times D}$ are added to $\vx_0$ as $\vz_0 = \vx_0 + \mE_{pos}$, where $\vz_0$ is the final input being fed to a sequence of transformer blocks.
	Each transformer block is composed of a multiheaded self-attention (MSA) and a MLP layer, together with Layernorm (LN) and skip connections, see Fig.~\ref{fig:adapter}(b). 
	The computation of a standard transformer block can be written as  
	\begin{equation}
		\vz'_{l} = \vz_{l-1} + \operatorname{MSA}(\operatorname{LN}(\vz_{l-1}))
	\end{equation}
	\begin{equation}
		\vz_{l} = \vz'_{l} + \operatorname{MLP}(\operatorname{LN}(\vz'_{l}))
	\end{equation}
	where $\vz_{l-1}$ and $\vz_{l}$ denotes the input and output of the $l$-th transformer block. Finally, the learned [class] token $\vx_{class}$ from the last transformer block is used as global visual representation and fed into a classification head to make the prediction.
	
	\textbf{Space-only and space-time models for video}.
	A video is a stack of frames with temporal structure. 
	Hence, video understanding requires the model to learn both good appearance representations in each frame (spatial modeling) and also infer the temporal structured information across frames (temporal modeling).
	In order to leverage an image transformer model for video tasks, one key thing is how to perform temporal modeling. 
	A simple baseline, termed space-only model, process each video frame independently by an image model.
	Given $\vx \in \R^{T \times H \times W \times C}$, where $T$ is the number of frames, space-only model will get $T$ [class] tokens where each [class] token stands for the representation of each frame. These $T$ [class] tokens will be averaged as a way of temporal modeling for final prediction. 
	In order to enhance the capability of temporal modeling, recent works \citep{timesformer,arnab2021vivit,zhang2021vidtr} introduce space-time model by adding new temporal modules to image models. 
	These models are now the top performers on most video action recognition benchmarks, however, their training costs are prohibitively high due to full finetuning. 
	Given the increasingly larger but more powerful pre-trained image models, in this work, we study how to efficiently adapt them for video action recognition. 

	\vspace{-10pt}
	\subsection{Spatial Adaptation}
	\label{subsec:spatial adapter}
	Since image pre-trained models have been trained on large-scale datasets and demonstrated strong transferability to downstream tasks, we believe they could achieve good spatial modeling in video action recognition with minimal finetuning. 
	
	Inspired by efficient finetuning techniques \citep{adapter, prompttuning, li-liang-2021-prefix, ben-zaken-etal-2022-bitfit} in NLP, we adopt Adapter \citep{adapter} due to its simplicity. 
	As shown in Fig.\ \ref{fig:adapter}(a), Adapter is a bottleneck architecture which consists of two fully connected (FC) layers and an activation layer in the middle.
	The first FC layer projects the input to a lower dimension and the second FC layer projects it back to the original dimension.
	To adapt the pre-trained spatial features to target video data, we add an Adapter after the self-attention layer as shown in Fig.\ \ref{fig:adapter}(c), which we term as spatial adaptation.
	During training, all the other layers of the transformer model are frozen while only the Adapters are updated. 
	In Table~\ref{tab:baseline}, we show that our spatial adaptation strategy achieves comparable performance with the full finetuned space-only baseline.
	This indicates that spatial adaptation helps the frozen image model to learn good spatial representations from video data. 
	However, the overall performance after spatial adaptation still has a large gap to a full finetuned video model because spatial adaptation alone lacks the ability to learn temporal information in videos.

	\subsection{Temporal Adaptation}
	\label{subsec:temporal module}
	
	To capture temporal information more effectively, previous methods usually incorporate new temporal modules to pre-trained image models because it is commonly believed that image models cannot infer temporal structured information in videos. 
	However, adding new temporal modules, either temporal attention \citep{timesformer,zhang2021vidtr} or temporal encoder/decoder \citep{arnab2021vivit,frozenclip}, will introduce sizable number of extra tunable parameters.
	In addition, these new modules require full finetuning, which is inefficient. 
	
	To address this problem, we present a new strategy: \textit{reuse the pre-trained self-attention layer in the image model to do temporal modeling}. 
	More specifically, we denote the original self-attention layer as S-MSA for spatial modeling, and the reused self-atentnion layer as T-MSA for temporal modeling. 
	As shown in Fig.\ \ref{fig:adapter}(d), we put T-MSA in front of S-MSA. 
	Now given the video patch embedding $\vz \in \R^{T \times (N+1) \times D}$,  we first reshape it into $\vz^T \in \R^{(N+1) \times T \times D}$, where $N = HW/P^2$ is the number of spatial patches and $T$ is the number of frames. 
	Then we feed $\vz^T$ into the T-MSA where it tries to learn the relationship among the $T$ frames. 
	Note that T-MSA and S-MSA are the same layer (i.e., pre-trained MSA in the image model) and kept frozen during model tuning, but just applied to different input dimensions. 
	This explicit operation helps our model with enhanced temporal modeling, while keeping the number of parameters fixed.
	In the end, similar to spatial adaptation, we add another Adapter after the reused temporal attention layer to adapt its features on video data, which we term as temporal adaptation.
	The structure of the Adapter is the same as in spatial adaptation but without the skip connection. 
	The reason is we want to initialize the adapted model to be close to the original model \citep{adapter}, thus we need to initialize the adapter to zero and remove the skip connection here to detach the effect of temporal adaptation at the beginning of training. 
	As seen in Table~\ref{tab:baseline}, temporal adaptation helps to close the gap to full finetuned video models while only introducing another lightweight Adapter into the transformer block.
	
	
	\subsection{Joint Adaptation}
	\label{subsec:joint module}
	
	Spatial and temporal adaptation are performed sequentially to different input dimensions with their individual purposes. 
	It would be desirable to jointly tune the  representations for spatiotemporal reasoning. 
	To this end, we further introduce an Adapter in parallel to the MLP layer, which we term as joint adaptation. 
	This Adapter has the same structure as the one in temporal adaptation.
	
	The final structure of a transformer block in our proposed AIM is presented in Fig.\ \ref{fig:adapter}(e). 
	The computation of the adapted block can be written as
	\begin{equation}
		\vz^T_{l} = \vz_{l-1} + \operatorname{Adapter}(\operatorname{T-MSA}(\operatorname{LN}(\vz_{l-1})))
	\end{equation}
	\begin{equation}
		\vz^S_{l} = \vz^T_{l} + \operatorname{Adapter}(\operatorname{S-MSA}(\operatorname{LN}(\vz^T_{l})))
	\end{equation}
	\begin{equation}
		\vz_{l} = \vz^S_{l} + \operatorname{MLP}(\operatorname{LN}(\vz^S_{l})) + s \cdot \operatorname{Adapter}(\operatorname{LN}(\vz^S_{l}))
	\end{equation}
	where $\vz^T_{l}$, $\vz^S_{l}$, $\vz_{l}$ denotes the temporal adapted, spatial adapted, and jointly adapted output in the  $l$-th transformer block. 
	Here, $s$ is a scaling factor to control the weight of the output from Adapter. 
	For the final prediction, we simply take the average of the [class] tokens of each input frame and feed it to the classification head.

	\section{Experiments}
	\label{sec:experiments}
	
	\textbf{Datasets.} We evaluate the proposed method on four widely adopted video action recognition benchmarks, Kinetics-400 (K400) \citep{kay2017kinetics}, Kinetics-700 (K700) \citep{k700}, Something-something-v2 (SSv2) \citep{goyal2017something} and Diving-48 \citep{diving48}. K400 contains around 240K training videos and 20K validation videos in 400 human action classes. The videos are all trimmed to around 10 seconds. K700 is an extended version of K400 which contains around 530K training videos and 34K validation videos in 700 classes. SSv2 contains 168.9K training videos and 24.7K validation videos in 174 classes. 
	SSv2 is more challenging because it requires stronger temporal modeling \citep{zhu2020videosurvey,sevilla2021only}.
	Diving-48 contains 15.9K training videos and 2K validation videos in 48 fine-grained diving actions. 
	It is designed to be unbiased towards static representations, which means a model cannot simply rely on the objects or background to determine the action. 
	\vspace{-2ex}
	\subsection{Effectiveness of Components}
	\label{subsec:baseline_exp}
	\vspace{-2ex}
	
	To demonstrate the effectiveness of our proposed components in Sec.\ \ref{sec:method}, we compare our method to three baselines. 
	The first baseline is a frozen space-only model. Recall in Sec.\ \ref{subsec:preliminary}, space-only model processes input frames independently and performs temporal average pooling in the end. We freeze the image backbone and only tune the classification head, which is also known as linear probing \citep{moco}. 
	The second baseline is a full finetuned space-only model. It should be able to learn spatial information from video data, but still has difficulties in capturing temporal information. 
	The third baseline is a full finetuned space-time video model, which should serves as oracle. 
	Here we choose TimeSformer \citep{timesformer} because we are based on the same ViT-B backbone and share a similar structure (\ie, divided space-time attention).
	
	In the experiments, we use the ViT-B/16 pre-trained on IN-21K as image backbone, and we compare the proposed method with the baselines on SSv2 \citep{goyal2017something} where temporal modeling is critical. 
	The results for three baselines are shown in Tab.\ \ref{tab:baseline} top. 
	We can see that the frozen space-only model only needs to  tune 0.1M parameters, but it also performs much worse than the full finetuned video model (15.1$\%$ vs 59.5$\%$). 
	Full finetuning the space-only model allows it to learn improved spatial representations from video data and largely improves the performance (15.1$\%$ $\rightarrow$ 36.2$\%$). 
	However, it also significantly increases the number of tunable parameters and still has a large gap from the full finetuned video model due to  lack of temporal modeling.
	The third baseline, full finetuned video model, achieves the highest accuracy due to its strong spatiotemporal reasoning capability, but the number of tunable parameters increases again to 121M.
	
	Our goal is to add a few tunable parameters to the frozen space-only model and close the gap to full finetuned video model. 
	As shown in Tab.\ \ref{tab:baseline} bottom, after spatial adaptation, the frozen space-only model achieves comparable performance with the full finetuned space-only model (36.7$\%$ vs 36.2$\%$), with significantly less number of tunable parameters (3.7M vs 86M).
	This means spatial adaptation is able to help frozen image models to achieve good spatial modeling on video data.
	In addition, adding temporal adaptation further boosts the performance to 61.2$\%$, which is even higher than the full finetuned video model. 
	This indicates that our temporal adaptation introduces strong temporal modeling to the space-only model. 
	Finally, joint adaptation is incorporated to tune the features for improved spatiotemporal reasoning, which is our method AIM. 
	We not only close the gap to full finetuned space-time video model but obtain higher accuracy (62$\%$ vs 59.5$\%$) with fewer number of tunable parameters (14.3M vs 86M).
	These results successfully validate the effectiveness of our proposed adaptation strategies. 
	
	Furthermore, our method could easily take advantage of stronger pre-trained image models and adapt them for video action recognition. 
	For example, simply switch the ViT-B/16 pre-trained on IN-21K to CLIP pre-trained, we obtain another accuracy boost (62.0$\%$ $\rightarrow$ 66.4$\%$)

	
	
	\begin{table}[t]
		\caption{Effectiveness of proposed components. We compare to three baselines on Something-something-v2 dataset. Spatial adaptation, temporal adaptation and joint adaptation gradually
			add spatiotemporal reasoning to the frozen image model. Views = \#frames $\times$ \#temporal $\times$ \#spatial.}
		\begin{center}
			\resizebox{\linewidth}{!}{
				\begin{tabular}{l|c|cc|cc|c}
					\hline
					Methods & Pretrain & Param (M) & \makecell{Tunable \\ Param (M)} & Top-1 & Top-5 & Views \\
					\hline
					Frozen space-only & IN-21K & 86 & 0.1 & 15.1 & 36.9 & 8$\times$1$\times$3 \\
					Finetuned space-only & IN-21K & 86 & 86 & 36.2 & 68.1 & 8$\times$1$\times$3 \\
					Finetuned space-time \citep{timesformer} & IN-21K & 121 & 121 & 59.5 & 85.6 & 8$\times$1$\times$3 \\
					\hline
					Frozen space-only + spatial adaptation & IN-21K & 89 & 3.7 & 36.7 & 68.3 & 8$\times$1$\times$3 \\
					\hspace{2.65cm} + temporal adaptation & IN-21K & 97 & 10.8 & 61.2 & 87.7 & 8$\times$1$\times$3 \\
					\hspace{2.65cm} + joint adaptation (AIM) & IN-21K & 100 & 14.3 & \textbf{62.0} & 87.9 & 8$\times$1$\times$3 \\
					\hline
					AIM & CLIP & 100 & 14.3 & \textbf{66.4} & 90.5 & 8$\times$1$\times$3 \\
					\hline
				\end{tabular}
			}
		\end{center}
		\label{tab:baseline}
		\vspace{-4ex}
	\end{table}
	
	\vspace{-1ex}
	\subsection{Comparisons to the State of the art}
	\vspace{-1ex}
	In this section, we compare the proposed method with state-of-the-art video models on four video action recognition benchmarks. For all the experiments, we use the ViT models pre-trained by CLIP \citep{clip}. 
	We mostly follow the training settings in \citet{liu2022videoswin}, and more implementation details can be found in Appendix.
	
	\subsubsection{Results on Kinetics-400 and Kinetics-700}
	\label{subsubsec:k400_results}

	Tab.\ \ref{tab:k400} presents the comparisons with state-of-the-art video models on K400 dataset. 
	First, we can see that with ViT-B/16 backbone, our method only needs to tune 11M parameters for competitive performance, which is much smaller than previous video models. 
	Taking input of 8 frames as an example, AIM ViT-B/16 achieves 83.9$\%$ top-1 accuracy while only requiring 606 GFLOPs. 
	When using 16 input frames, our method even outperforms MTV-L \citep{yan2022multiview}, which requires more than 10$\times$ computations (1214 vs 18050 GFLOPs).
	When switching to larger backbone ViT-L/14, we achieve the highest accuracy $87.5\%$ on K400 dataset, with 38M tunable parameters. 
	
	Note that several works also leverage CLIP pre-trained models to do video action recognition.
	However, ActionCLIP \citep{wang2021actionclip} and X-CLIP \citep{xclip} are multimodal methods which require additional text branch and tune the whole model end-to-end. PromptCLIP \citep{promptclip} applies prompt tuning \citep{prompttuning} to CLIP  and adds several temporal blocks for temporal modeling.
	EVL \citep{frozenclip} introduces a new decoder branch to learn temporal information. However, AIM simply re-uses image pre-trained self-attention for temporal modeling. This makes AIM much simpler than previous methods, yet achieving better performance at much less tunable parameters. The simplicity also makes AIM much easier to adapt to different model architectures (single modal or multi-modal models). But previous methods such as ActionCLIP/X-CLIP/PromptCLIP cannot leverage pure image backbone because they need an additional text branch.
	
	Furthermore, we evaluate our method on K700 dataset in Tab.~\ref{tab:k700}. We can see that AIM ViT-B/16 with 11M tunable parameters is able to outperform MTV-L (875M) and MViTv2-B (51M). And AIM ViT-L/14 (38M) achieves comparable performance with MaskFeat (218M) \citep{maskfeat}. Note that MaskFeat uses larger input resolution (312 vs 224) and more input frames (40 vs 32) than us.
	This again justifies the effectiveness of our efficient adaptation pipeline. 

	\begin{table}[t]
		\caption{Comparison to state-of-the-art on Kinetics-400. Views = \#frames $\times$ \#temporal $\times$ \#spatial.}
		\vspace{-2ex}
		\begin{center}
			\resizebox{\linewidth}{!}{
				\begin{tabular}{l|c|ccc|cc|c}
					\hline
					Methods & Pretrain & GFLOPs & \makecell{Param \\ (M)} & \makecell{Tunable \\ Param (M)} & Top-1 & Top-5 & Views \\
					\hline
					MViT-B \citep{mvit} & - & 4095 & 37 & 37 & 81.2 & 95.1 & 64$\times$3$\times$3 \\
					UniFormer-B \citep{li2021uniformer} & IN-1K & 3108 & 50 & 50 & 83.0 & 95.4 & 32$\times$4$\times$3 \\
					TimeSformer-L \citep{timesformer} & IN-21K & 7140 & 121 & 121 & 80.7 & 94.7 & 64$\times$1$\times$3 \\
					ViViT-L/16$\times$2 FE \citep{arnab2021vivit} & IN-21K & 3980 & 311 & 311 & 80.6 & 92.7 & 32$\times$1$\times$1 \\
					VideoSwin-L \citep{liu2022videoswin} & IN-21K & 7248 & 197 & 197 & 83.1 & 95.9 & 32$\times$4$\times$3 \\
					MViTv2-L ($312\uparrow$) \citep{mvitv2} & IN-21K & 42420 & 218 & 218 & 86.1 & 97.0 & 32$\times$3$\times$5 \\
					MTV-L \citep{yan2022multiview} & JFT & 18050 & 876 & 876 & 84.3 & 96.3 & 32$\times$4$\times$3 \\
					TokenLearner-L/10 \citep{ryoo2021tokenlearner} & JFT & 48912 & 450 & 450 & 85.4 & 96.3 & 64$\times$4$\times$3 \\
					PromptCLIP A7 \citep{promptclip} & CLIP & - & - & - & 76.8 & 93.5 & 16$\times$5$\times$1 \\
					ActionCLIP \citep{wang2021actionclip} & CLIP & 16890 & 142 & 142 & 83.8 & 97.1 & 32$\times$10$\times$3 \\
					X-CLIP-L/14 \citep{xclip} & CLIP & 7890 & 420 & 420 & 87.1 & 97.6 & 8$\times$4$\times$3 \\
					EVL ViT-L/14 \citep{frozenclip} & CLIP & 8088 & 368 & 59 & 87.3 & - & 32$\times$3$\times$1 \\
					\hline
					AIM ViT-B/16 & CLIP & 606 & 97 & 11 & 83.9 & 96.3 & 8$\times$3$\times$1 \\
					AIM ViT-B/16 & CLIP & 1214 & 97 & 11 & 84.5 & 96.6 & 16$\times$3$\times$1 \\
					AIM ViT-B/16 & CLIP & 2428 & 97 & 11 & 84.7 & 96.7 & 32$\times$3$\times$1 \\
					
					AIM ViT-L/14 & CLIP & 2802 & 341 & 38 & 86.8 & 97.2 & 8$\times$3$\times$1 \\
					AIM ViT-L/14 & CLIP & 5604 & 341 & 38 & 87.3 & 97.6 & 16$\times$3$\times$1 \\
					AIM ViT-L/14 & CLIP & 11208 & 341 & 38 & \textbf{87.5} & \textbf{97.7} & 32$\times$3$\times$1 \\
					\hline
				\end{tabular}
			}
		\end{center}
		\label{tab:k400}
		\vspace{-2ex}
	\end{table}
	
	\subsubsection{Results on Something-Something-v2}
	Tab. \ref{tab:ssv2} presents the performance comparisons on SSv2. 
	Based on CLIP ViT-L/14, our method achieves competitive or better performance than most  prior arts.
	In terms of fair comparison to EVL, which also uses CLIP pre-trained image encoder, we achieve significantly higher accuracy ($70.6\%$ $>$ $66.7\%$), while introducing  less tunable parameters (50M $<$ 175M). 
	Note that to introduce temporal modeling into image model, EVL adds 12 layers of decoder blocks, while our method  simply reuse image pre-trained self-attention layers to achieve stronger temporal modeling . 
	
	However, our method falls behind some full finetuned video models \citep{girdhar2022omnivore,mvitv2,li2021uniformer}. 
	One reason is that SSv2 is a ``temporal-heavy" dataset \citep{sevilla2021only}, which requires model to really understand the temporal evolution within a video.
	In order to obtain high accuracy, most previous video models are first pre-trained on some video datasets (such as K400/K600) to learn good spatiotemporal representations, then finetuned on SSv2. 
	But our method still starts from the image pre-trained model.
	Another reason is that simply reusing the image pre-trained self-attention for temporal modeling may not be able to fully capture the complicated temporal information in SSv2 videos. 
	This suggests that we need to conduct more temporal adaptation for these challenging ``temporal-heavy" datasets.

	\begin{table}[t]
		\caption{Comparison to state-of-the-art on Something-Something-v2. K400$^\dagger$/K600$^\dagger$ indicates the model is pre-trained on both IN-21K and K400/K600.}
		\vspace{-1ex}
		\begin{center}
			\resizebox{\linewidth}{!}{
				\begin{tabular}{l|c|ccc|cc|c}
					\hline
					Methods & Pretrain & GFLOPs & \makecell{Param \\ (M)} & \makecell{Tunable \\ Param (M)} & Top-1 & Top-5 & Views \\
					\hline
					TimeSformer-L \citep{timesformer} & IN-21K & 7140 & 121 & 121 & 62.4 & - & 64$\times$1$\times$3 \\
					MTV-B \citep{yan2022multiview} & IN-21K & 4790 & 310 & 310 & 67.6 & 90.4 & 32$\times$4$\times$3 \\
					MViT-B \citep{mvit} & K400 & 510 & 37 & 37 & 67.1 & 90.8 & 32$\times$1$\times$3 \\
					MViTv2-B \citep{mvitv2} & K400 & 675 & 51 & 51 & 70.5 & 92.7 & 40$\times$1$\times$3 \\
					ViViT-L/16$\times$2 \citep{arnab2021vivit} & K400$^\dagger$ & 11892 & 311 & 311 & 65.4 & 89.8 & 16$\times$4$\times$3 \\
					VideoSwin-B \citep{liu2022videoswin} & K400$^\dagger$ & 963 & 89 & 89 & 69.6 & 92.7 & 32$\times$1$\times$1 \\
					Omnivore \citep{girdhar2022omnivore} & K400$^\dagger$ & - & - & - & 71.4 & 93.5 & 32$\times$1$\times$3 \\
					MViTv2-L ($312 \uparrow$) \citep{mvitv2} & K400$^\dagger$ & 8484 & 213 & 213 & \textbf{73.3} & \textbf{94.1} & 32$\times$1$\times$3 \\
					
					UniFomer-B \citep{li2021uniformer} & K600$^\dagger$ & 777 & 50 & 50 & 71.2 & 92.8 & 32$\times$1$\times$3 \\
					CoVeR \citep{cover} & JFT-3B & - & - & - & 70.9 & - & - \\
					EVL ViT-B/16 \citep{frozenclip} & CLIP & 2047 & 182 & 86 & 62.4 & - & 32$\times$1$\times$3 \\
					EVL ViT-L/14 \cite{frozenclip} & CLIP & 9641 & 484 & 175 & 66.7 & - & 32$\times$1$\times$3 \\
					\hline
					AIM ViT-B/16 & CLIP & 624 & 100 & 14 & 66.4 & 90.5 & 8$\times$1$\times$3 \\
					AIM ViT-B/16 & CLIP & 1248 & 100 & 14 & 68.1 & 91.8 & 16$\times$1$\times$3 \\
					AIM ViT-B/16 & CLIP & 2496 & 100 & 14 & 69.1 & 92.2 & 32$\times$1$\times$3 \\
					AIM ViT-L/14 & CLIP & 2877 & 354 & 50 & 67.6 & 91.6 & 8$\times$1$\times$3 \\
					AIM ViT-L/14 & CLIP & 5754 & 354 & 50 & 69.4 & 92.3 & 16$\times$1$\times$3 \\
					AIM ViT-L/14 & CLIP & 11508 & 354 & 50 & 70.6 & 92.7 & 32$\times$1$\times$3 \\
					\hline
				\end{tabular}
			}
		\end{center}
		\label{tab:ssv2}
		\vspace{-4ex}
	\end{table}
	
	\subsubsection{Results on Diving-48}
	A diving class in Diving-48 \citep{diving48} is defined by the combination of takeoff, movements in flight and entry, thus it requires the model to differentiate such fine-grained actions. 
	As shown in Tab.\ \ref{tab:diving48}, our method with 11M tunable parameters outperforms all prior methods. AIM ViT-L/14 further improves the top-1 accuracy to 90.6\%.
	Comparing to ORViT \citep{orvit}, despite they leverage additional object tracking model, our method still outperforms it with much less tunable parameters. 
	This suggests that efficient finetuning can handle fine-grained action recognition.

	\begin{figure}
		\begin{minipage}[t]{0.49\linewidth}
			\centering
			\tabcaption{Comparisons on Kinetics-700.}
			\resizebox{\linewidth}{!}{
				\begin{tabular}{l|c|c|c}
					\hline
					Method & Pretrain & \makecell{Tunable \\ Param} & Top-1 \\
					\hline
					VidTR-L \citep{zhang2021vidtr} & IN-21K & 91 & 70.2  \\
					MTV-L \citep{yan2022multiview} & IN-21K & 876 & 75.2 \\
					MViTv2-B \citep{mvitv2} & - & 51 & 76.6 \\
					MViTv2-L ($40\times312\uparrow$) \citep{mvitv2} & IN-21K & 218 & 79.4 \\
					MaskFeat ($40\times312\uparrow$) \citep{maskfeat} & K700 & 218 & \textbf{80.4} \\
					\hline
					AIM ViT-B/16 & CLIP & 11 & 76.9 \\
					AIM ViT-L/14 & CLIP & 38 & \textbf{80.4} \\
					\hline
				\end{tabular}
			}
			\label{tab:k700}
		\end{minipage}%
		\hspace{4pt}
		\begin{minipage}[t]{0.49\linewidth}
			\centering
			\tabcaption{Comparisons on Diving-48.}
			\resizebox{\linewidth}{!}{
				\begin{tabular}{l|c|c|c}
					\hline
					Method & Pretrain & \makecell{Tunable \\ Param} & Top-1 \\
					\hline
					TimeSformer-L \citep{timesformer} & IN-21K & 121 & 81.0  \\
					VideoSwin-B \citep{liu2022videoswin} & IN-21K & 88 & 81.9 \\
					BEVT \citep{wang2022bevt} & K400$^\dagger$ & 88 & 86.7 \\
					SIFAR-B-14 \citep{sifar} & IN-21K & 87 & 87.3 \\
					\color{gray} ORViT \citep{orvit} & \color{gray} IN-21K &  \color{gray} 160 & \color{gray} 88.0 \\
					\hline
					AIM ViT-B/16 & CLIP & 11 & 88.9 \\
					AIM ViT-L/14 & CLIP & 38 & \textbf{90.6} \\
					\hline
				\end{tabular}
			}
			\label{tab:diving48}
		\end{minipage}%
		\vspace{-2ex}
	\end{figure}
	\vspace{-1ex}
	\section{Discussion}
	\vspace{-1ex}
	\begin{wraptable}{r}{0.5\textwidth}
		\vspace{-15pt}
		\caption{Performance of using different pre-trained models on K400.}
		\vspace{-10pt}
		\begin{center}
			\resizebox{0.5\textwidth}{!}{
				\begin{tabular}{l|cc|ccc|c}
					\hline
					Model & Backbone & Pretrain & \makecell{Tunable \\ Param (M)} & \makecell{Mem \\ (G)} & \makecell{Time \\ (H)} & Top-1  \\
					\hline
					TimeSformer & ViT-B & IN-21K & 121 & 10 & 20 & 78.5 \\
					AIM & ViT-B &  IN-21K & 11 & 7 & 15 & 78.8 \\
					TimeSformer & ViT-B & CLIP & 121 & 10 & 20 & 82.0 \\
					AIM & ViT-B & CLIP & 11 & 7 & 15 & 83.9 \\
					\hline
					VideoSwin-B & Swin-B & IN-21K & 88 & 18 & 64 & 82.7 \\
					AIM & Swin-B & IN-21K & 9.2 & 9 & 37 & 82.1 \\
					\hline
				\end{tabular}%
			}
		\end{center}
		\label{tab:pretrain}
		\vspace{-8pt}
	\end{wraptable}
	\textbf{Different Pre-trained Models.} Here we demonstrate the effectiveness of AIM on different pre-trained models. In Table \ref{tab:pretrain}, we first show AIM based on ViT-B backbone. We compare AIM to TimeSformer because we use the same backbone (ViT-B) and have a similar structure (i.e., both using divided space-time attention).
	As can be seen, AIM achieves better performance than full finetuned TimeSformer under both IN-21K and CLIP pre-trained weights. Then we apply AIM to Swin-B backbone and compare it to VideoSwin when we both use Swin-B and IN-21K pre-training. Similarly, AIM achieves comparable performance with full finetuned VideoSwin.
	
\begin{figure}
		\begin{minipage}[t]{0.5\linewidth}
			\vspace{3pt}
			\centering
			\includegraphics[width=\linewidth]{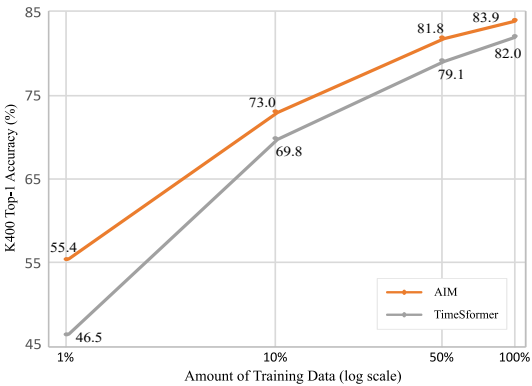}
			\vspace{-15pt}
			\figcaption{Data efficiency comparison. AIM outperforms full finetuned TimeSformer under all scenarios, especially in low data regime.}
			\label{fig:dataefficiency}
		\end{minipage}
		\hspace{0.3cm}
		\begin{minipage}[t]{0.48\linewidth}
			\vspace{-5pt}
			\centering
			\tabcaption{Effect of position of Adapters. Skip means adding Adapters every two blocks.}
			\vspace{5pt}
			\resizebox{0.7\linewidth}{!}{
				\begin{tabular}{l|c|c}
					\hline
					Position & \makecell{Tunable \\ Param (M)} & Top-1 \\
					\hline
					Bottom 6 & 5.6 & 80.7 \\
					Top 6 & 5.6 & 83.3 \\
					Skip & 5.6 & 83.2 \\
					All & 11 & \textbf{83.9} \\
					\hline
				\end{tabular}
			}
			\label{tab:positionadapter}
			\tabcaption{Effect of bottleneck ratio of Adapters.}
			\vspace{5pt}
			\resizebox{0.7\linewidth}{!}{
				\begin{tabular}{c|c|c}
					\hline
					Ratio & \makecell{Tunable \\ Param (M)} & Top-1 \\
					\hline
					0.0625 & 3 & 83.3  \\
					0.125 & 5.6 & 83.4 \\
					0.25 & 11 & \textbf{83.9} \\
					0.5 & 21 & 83.8 \\
					\hline
				\end{tabular}
			}
			\label{tab:bottleneckratio}
		\end{minipage}%
		\vspace{-3ex}
	\end{figure}

	\textbf{Data Efficiency.} One advantage of our efficient tuning paradigm is that we can keep the well pre-trained image representations intact. In the scenario where downtream data is insufficient, our method will be less prone to over-fitting compared to full finetuning. 
	In Fig.\ \ref{fig:dataefficiency}, we compare AIM with full finetuned TimeSformer under different amounts of training data on K400. For fair comparison, both AIM and TimeSformer use CLIP pre-trained ViT-B/16 as backbone. 
	We can observe that under all scenarios, our method AIM outperforms full finetuned TimeSformer. 
	In particular, when the amount of data becomes less, the advantage of AIM becomes larger. 
	For example, when there is only $1\%$ of training data, we  outperform TimeSformer by a significant margin of $8.9\%$.
	
	\textbf{Training Cost.} Tab.\ \ref{tab:pretrain} also shows the training time (hours) and memory cost (GB) of our method and full finetuning on different backbones. All metrics are measured on 8 Tesla V100 GPUs. Compared to TimeSformer, we reduce the memory cost by 30\% and training time by 25\%. Compared to VideoSwin, we reduce the memory cost by 50\% and training time by 42\%.
	
	\textbf{Position of Adapters.} By default, we add Adapters to every ViT block (12 blocks in total). Here we study the effect of adding Adapters in different layers. We add Adapters to the bottom 6 blocks (close to the input), top 6 blocks (close to the output) and one every two blocks. All these variants have the same number of tunable parameters.
	As can be seen in Tab.\ \ref{tab:positionadapter},  adding Adapters to the bottom 6 blocks yields much worse performance than others. We hypothesize that the shallow layers learn generic representations which do not need much adaptation, while deeper layers learn task-specific features like temporal information thus feature adaptation is important. 
	Adding Adapters to the top 6 blocks achieves comparable performance with adding to all blocks while saving half of the parameters. 
	This could serve as a good candidate when training resources are more limited. 
	
	\textbf{Bottleneck Ratio of Adapters.} 
	By tuning the bottleneck ratio of Adapters, we can easily control the number of tunable parameters. 
	Here we study how the bottleneck ratio of Adapters affects the final performance. 
	The results in Tab.\ \ref{tab:bottleneckratio} reveal that a larger bottleneck ratio tends to achieve better performance, but it will also introduce more tunable parameters. 
	The performance plateaus after bottleneck ratio goes beyond 0.25. Note that a small ratio of 0.0625 could still achieve 83.3\% top-1 accuracy on K400, which is competitive among state-of-the-art video models in Tab.\ \ref{tab:k400} while introducing only 3M tunable parameters.
	
	\vspace{-2ex}
	\section{Conclusion}
	\vspace{-2ex}
	In this work, we propose a new way to efficiently transfer pre-trained image models for video action recognition. 
	We introduce spatial adaptation, temporal adaptation and joint adaptation to gradually add spatiotemporal reasoning to an image model.
	Since only newly added Adapters are updated, our training cost is substantially lower than other full finetuned video models. 
	Yet we achieve comparable or even better performance than prior arts on four benchmarks. 
	Our method is simple and generally applicable, which has the potential to leverage more powerful image foundation models in the future. 
	Despite all the benefits, one limitation is that our simple strategy of reusing spatial attention for temporal modeling might not be strong enough for temporally challenging videos.
	Since video temporal modeling can be viewed as a form of sequence modeling, we might be able to reuse pre-trained weights from text or audio models instead of image models in the future. 

	


	\bibliography{iclr2023_conference}

\begin{thebibliography}{76}
\providecommand{\natexlab}[1]{#1}
\providecommand{\url}[1]{\texttt{#1}}
\expandafter\ifx\csname urlstyle\endcsname\relax
  \providecommand{\doi}[1]{doi: #1}\else
  \providecommand{\doi}{doi: \begingroup \urlstyle{rm}\Url}\fi

\bibitem[Arnab et~al.(2021)Arnab, Dehghani, Heigold, Sun, Lu{\v{c}}i{\'c}, and
  Schmid]{arnab2021vivit}
Anurag Arnab, Mostafa Dehghani, Georg Heigold, Chen Sun, Mario Lu{\v{c}}i{\'c},
  and Cordelia Schmid.
\newblock Vivit: A video vision transformer.
\newblock In \emph{Proceedings of the IEEE/CVF International Conference on
  Computer Vision}, pp.\  6836--6846, 2021.

\bibitem[Bahng et~al.(2022)Bahng, Jahanian, Sankaranarayanan, and
  Isola]{visualpixelprompt}
Hyojin Bahng, Ali Jahanian, Swami Sankaranarayanan, and Phillip Isola.
\newblock Visual prompting: Modifying pixel space to adapt pre-trained models.
\newblock \emph{arXiv preprint arXiv:2203.17274}, 2022.

\bibitem[Bao et~al.(2021)Bao, Dong, Piao, and Wei]{beit}
Hangbo Bao, Li~Dong, Songhao Piao, and Furu Wei.
\newblock {BEiT: BERT Pre-Training of Image Transformers}.
\newblock \emph{arXiv preprint arXiv:2106.08254}, 2021.

\bibitem[Ben~Zaken et~al.(2022)Ben~Zaken, Goldberg, and
  Ravfogel]{ben-zaken-etal-2022-bitfit}
Elad Ben~Zaken, Yoav Goldberg, and Shauli Ravfogel.
\newblock {B}it{F}it: Simple parameter-efficient fine-tuning for
  transformer-based masked language-models.
\newblock In \emph{Proceedings of the 60th Annual Meeting of the Association
  for Computational Linguistics (Volume 2: Short Papers)}, pp.\  1--9, Dublin,
  Ireland, May 2022. Association for Computational Linguistics.
\newblock \doi{10.18653/v1/2022.acl-short.1}.
\newblock URL \url{https://aclanthology.org/2022.acl-short.1}.

\bibitem[Bertasius et~al.(2021)Bertasius, Wang, and Torresani]{timesformer}
Gedas Bertasius, Heng Wang, and Lorenzo Torresani.
\newblock Is space-time attention all you need for video understanding?
\newblock In \emph{Proceedings of the International Conference on Machine
  Learning (ICML)}, July 2021.

\bibitem[Brown et~al.(2020)Brown, Mann, Ryder, Subbiah, Kaplan, Dhariwal,
  Neelakantan, Shyam, Sastry, Askell, et~al.]{gpt}
Tom Brown, Benjamin Mann, Nick Ryder, Melanie Subbiah, Jared~D Kaplan, Prafulla
  Dhariwal, Arvind Neelakantan, Pranav Shyam, Girish Sastry, Amanda Askell,
  et~al.
\newblock Language models are few-shot learners.
\newblock \emph{Advances in Neural Information Processing Systems},
  33:\penalty0 1877--1901, 2020.

\bibitem[Carreira \& Zisserman(2017)Carreira and Zisserman]{i3d}
Joao Carreira and Andrew Zisserman.
\newblock Quo vadis, action recognition? a new model and the kinetics dataset.
\newblock In \emph{proceedings of the IEEE Conference on Computer Vision and
  Pattern Recognition}, pp.\  6299--6308, 2017.

\bibitem[Carreira et~al.(2019)Carreira, Noland, Hillier, and Zisserman]{k700}
Joao Carreira, Eric Noland, Chloe Hillier, and Andrew Zisserman.
\newblock A short note on the kinetics-700 human action dataset.
\newblock \emph{arXiv preprint arXiv:1907.06987}, 2019.

\bibitem[Chen et~al.(2022)Chen, Ge, Tong, Wang, Song, Wang, and
  Luo]{chen2022adaptformer}
Shoufa Chen, Chongjian Ge, Zhan Tong, Jiangliu Wang, Yibing Song, Jue Wang, and
  Ping Luo.
\newblock Adaptformer: Adapting vision transformers for scalable visual
  recognition.
\newblock \emph{arXiv preprint arXiv:2205.13535}, 2022.

\bibitem[Chen et~al.(2021)Chen, Xie, and He]{mocov3}
Xinlei Chen, Saining Xie, and Kaiming He.
\newblock An empirical study of training self-supervised vision transformers.
\newblock In \emph{Proceedings of the IEEE/CVF International Conference on
  Computer Vision}, pp.\  9640--9649, 2021.

\bibitem[Cubuk et~al.(2020)Cubuk, Zoph, Shlens, and Le]{cubuk2020randaugment}
Ekin~D Cubuk, Barret Zoph, Jonathon Shlens, and Quoc~V Le.
\newblock Randaugment: Practical automated data augmentation with a reduced
  search space.
\newblock In \emph{Proceedings of the IEEE/CVF Conference on Computer Vision
  and Pattern Recognition workshops}, pp.\  702--703, 2020.

\bibitem[Deng et~al.(2009)Deng, Dong, Socher, Li, Li, and
  Fei-Fei]{deng2009imagenet}
Jia Deng, Wei Dong, Richard Socher, Li-Jia Li, Kai Li, and Li~Fei-Fei.
\newblock Imagenet: A large-scale hierarchical image database.
\newblock In \emph{2009 IEEE Conference on Computer Vision and Pattern
  Recognition}, pp.\  248--255. IEEE, 2009.

\bibitem[Devlin et~al.(2018)Devlin, Chang, Lee, and Toutanova]{devlin2018bert}
Jacob Devlin, Ming-Wei Chang, Kenton Lee, and Kristina Toutanova.
\newblock Bert: Pre-training of deep bidirectional transformers for language
  understanding.
\newblock \emph{arXiv preprint arXiv:1810.04805}, 2018.

\bibitem[Dong et~al.(2022)Dong, Bao, Chen, Zhang, Yu, Yuan, Chen, and
  Guo]{dong2022cswin}
Xiaoyi Dong, Jianmin Bao, Dongdong Chen, Weiming Zhang, Nenghai Yu, Lu~Yuan,
  Dong Chen, and Baining Guo.
\newblock Cswin transformer: A general vision transformer backbone with
  cross-shaped windows.
\newblock In \emph{Proceedings of the IEEE/CVF Conference on Computer Vision
  and Pattern Recognition}, pp.\  12124--12134, 2022.

\bibitem[Dosovitskiy et~al.(2020)Dosovitskiy, Beyer, Kolesnikov, Weissenborn,
  Zhai, Unterthiner, Dehghani, Minderer, Heigold, Gelly, et~al.]{vit}
Alexey Dosovitskiy, Lucas Beyer, Alexander Kolesnikov, Dirk Weissenborn,
  Xiaohua Zhai, Thomas Unterthiner, Mostafa Dehghani, Matthias Minderer, Georg
  Heigold, Sylvain Gelly, et~al.
\newblock An image is worth 16x16 words: Transformers for image recognition at
  scale.
\newblock In \emph{International Conference on Learning Representations}, 2020.

\bibitem[Fan et~al.(2021)Fan, Xiong, Mangalam, Li, Yan, Malik, and
  Feichtenhofer]{mvit}
Haoqi Fan, Bo~Xiong, Karttikeya Mangalam, Yanghao Li, Zhicheng Yan, Jitendra
  Malik, and Christoph Feichtenhofer.
\newblock Multiscale vision transformers.
\newblock In \emph{Proceedings of the IEEE/CVF International Conference on
  Computer Vision}, pp.\  6824--6835, 2021.

\bibitem[Fan et~al.(2022)Fan, Chen, and Panda]{sifar}
Quanfu Fan, Chun-Fu Chen, and Rameswar Panda.
\newblock Can an image classifier suffice for action recognition?
\newblock In \emph{International Conference on Learning Representations}, 2022.
\newblock URL \url{https://openreview.net/forum?id=qhkFX-HLuHV}.

\bibitem[Feichtenhofer et~al.(2019)Feichtenhofer, Fan, Malik, and
  He]{feichtenhofer2019slowfast}
Christoph Feichtenhofer, Haoqi Fan, Jitendra Malik, and Kaiming He.
\newblock Slowfast networks for video recognition.
\newblock In \emph{Proceedings of the IEEE/CVF International Conference on
  Computer Vision}, pp.\  6202--6211, 2019.

\bibitem[Feichtenhofer et~al.(2022)Feichtenhofer, Fan, Li, and
  He]{feichtenhofer2022videomae}
Christoph Feichtenhofer, Haoqi Fan, Yanghao Li, and Kaiming He.
\newblock Masked autoencoders as spatiotemporal learners.
\newblock \emph{arXiv preprint arXiv:2205.09113}, 2022.

\bibitem[Gao et~al.(2022)Gao, Shi, Zhu, Wang, Tang, Zhou, Li, and
  Metaxas]{gao2022dept}
Yunhe Gao, Xingjian Shi, Yi~Zhu, Hao Wang, Zhiqiang Tang, Xiong Zhou, Mu~Li,
  and Dimitris~N. Metaxas.
\newblock {Visual Prompt Tuning for Test-time Domain Adaptation}.
\newblock \emph{arXiv preprint arXiv:2210.04831}, 2022.

\bibitem[Girdhar et~al.(2022)Girdhar, Singh, Ravi, van~der Maaten, Joulin, and
  Misra]{girdhar2022omnivore}
Rohit Girdhar, Mannat Singh, Nikhila Ravi, Laurens van~der Maaten, Armand
  Joulin, and Ishan Misra.
\newblock Omnivore: A single model for many visual modalities.
\newblock In \emph{Proceedings of the IEEE/CVF Conference on Computer Vision
  and Pattern Recognition}, pp.\  16102--16112, 2022.

\bibitem[Goyal et~al.(2017)Goyal, Ebrahimi~Kahou, Michalski, Materzynska,
  Westphal, Kim, Haenel, Fruend, Yianilos, Mueller-Freitag,
  et~al.]{goyal2017something}
Raghav Goyal, Samira Ebrahimi~Kahou, Vincent Michalski, Joanna Materzynska,
  Susanne Westphal, Heuna Kim, Valentin Haenel, Ingo Fruend, Peter Yianilos,
  Moritz Mueller-Freitag, et~al.
\newblock The" something something" video database for learning and evaluating
  visual common sense.
\newblock In \emph{Proceedings of the IEEE International Conference on Computer
  Vision}, pp.\  5842--5850, 2017.

\bibitem[He et~al.(2022{\natexlab{a}})He, Zhou, Ma, Berg-Kirkpatrick, and
  Neubig]{unifiedadapter}
Junxian He, Chunting Zhou, Xuezhe Ma, Taylor Berg-Kirkpatrick, and Graham
  Neubig.
\newblock Towards a unified view of parameter-efficient transfer learning.
\newblock In \emph{International Conference on Learning Representations},
  2022{\natexlab{a}}.
\newblock URL \url{https://openreview.net/forum?id=0RDcd5Axok}.

\bibitem[He et~al.(2020)He, Fan, Wu, Xie, and Girshick]{moco}
Kaiming He, Haoqi Fan, Yuxin Wu, Saining Xie, and Ross Girshick.
\newblock Momentum contrast for unsupervised visual representation learning.
\newblock In \emph{Proceedings of the IEEE/CVF Conference on Computer Vision
  and Pattern Recognition}, pp.\  9729--9738, 2020.

\bibitem[He et~al.(2022{\natexlab{b}})He, Chen, Xie, Li, Doll{\'a}r, and
  Girshick]{mae}
Kaiming He, Xinlei Chen, Saining Xie, Yanghao Li, Piotr Doll{\'a}r, and Ross
  Girshick.
\newblock Masked autoencoders are scalable vision learners.
\newblock In \emph{Proceedings of the IEEE/CVF Conference on Computer Vision
  and Pattern Recognition}, pp.\  16000--16009, 2022{\natexlab{b}}.

\bibitem[Herzig et~al.(2022)Herzig, Ben-Avraham, Mangalam, Bar, Chechik,
  Rohrbach, Darrell, and Globerson]{orvit}
Roei Herzig, Elad Ben-Avraham, Karttikeya Mangalam, Amir Bar, Gal Chechik, Anna
  Rohrbach, Trevor Darrell, and Amir Globerson.
\newblock Object-region video transformers.
\newblock In \emph{Proceedings of the IEEE/CVF Conference on Computer Vision
  and Pattern Recognition}, pp.\  3148--3159, 2022.

\bibitem[Houlsby et~al.(2019)Houlsby, Giurgiu, Jastrzebski, Morrone,
  De~Laroussilhe, Gesmundo, Attariyan, and Gelly]{adapter}
Neil Houlsby, Andrei Giurgiu, Stanislaw Jastrzebski, Bruna Morrone, Quentin
  De~Laroussilhe, Andrea Gesmundo, Mona Attariyan, and Sylvain Gelly.
\newblock Parameter-efficient transfer learning for nlp.
\newblock In \emph{International Conference on Machine Learning}, pp.\
  2790--2799. PMLR, 2019.

\bibitem[Hu et~al.(2022)Hu, yelong shen, Wallis, Allen-Zhu, Li, Wang, Wang, and
  Chen]{hu2022lora}
Edward~J Hu, yelong shen, Phillip Wallis, Zeyuan Allen-Zhu, Yuanzhi Li, Shean
  Wang, Lu~Wang, and Weizhu Chen.
\newblock Lo{RA}: Low-rank adaptation of large language models.
\newblock In \emph{International Conference on Learning Representations}, 2022.
\newblock URL \url{https://openreview.net/forum?id=nZeVKeeFYf9}.

\bibitem[Jia et~al.(2021)Jia, Yang, Xia, Chen, Parekh, Pham, Le, Sung, Li, and
  Duerig]{align}
Chao Jia, Yinfei Yang, Ye~Xia, Yi-Ting Chen, Zarana Parekh, Hieu Pham, Quoc Le,
  Yun-Hsuan Sung, Zhen Li, and Tom Duerig.
\newblock Scaling up visual and vision-language representation learning with
  noisy text supervision.
\newblock In \emph{International Conference on Machine Learning}, pp.\
  4904--4916. PMLR, 2021.

\bibitem[Jia et~al.(2022)Jia, Tang, Chen, Cardie, Belongie, Hariharan, and
  Lim]{vpt}
Menglin Jia, Luming Tang, Bor-Chun Chen, Claire Cardie, Serge Belongie, Bharath
  Hariharan, and Ser-Nam Lim.
\newblock Visual prompt tuning.
\newblock \emph{arXiv preprint arXiv:2203.12119}, 2022.

\bibitem[Jie \& Deng(2022)Jie and Deng]{convadapter}
Shibo Jie and Zhi-Hong Deng.
\newblock Convolutional bypasses are better vision transformer adapters.
\newblock \emph{arXiv preprint arXiv:2207.07039}, 2022.

\bibitem[Ju et~al.(2021)Ju, Han, Zheng, Zhang, and Xie]{promptclip}
Chen Ju, Tengda Han, Kunhao Zheng, Ya~Zhang, and Weidi Xie.
\newblock Prompting visual-language models for efficient video understanding.
\newblock \emph{arXiv preprint arXiv:2112.04478}, 2021.

\bibitem[Kay et~al.(2017)Kay, Carreira, Simonyan, Zhang, Hillier,
  Vijayanarasimhan, Viola, Green, Back, Natsev, et~al.]{kay2017kinetics}
Will Kay, Joao Carreira, Karen Simonyan, Brian Zhang, Chloe Hillier, Sudheendra
  Vijayanarasimhan, Fabio Viola, Tim Green, Trevor Back, Paul Natsev, et~al.
\newblock The kinetics human action video dataset.
\newblock \emph{arXiv preprint arXiv:1705.06950}, 2017.

\bibitem[Kingma \& Ba(2014)Kingma and Ba]{kingma2014adam}
Diederik~P Kingma and Jimmy Ba.
\newblock Adam: A method for stochastic optimization.
\newblock \emph{arXiv preprint arXiv:1412.6980}, 2014.

\bibitem[Kuang et~al.(2021)Kuang, Zhu, Zhang, Li, Tighe, Schwertfeger,
  Stachniss, and Li]{kuang2021video}
Haofei Kuang, Yi~Zhu, Zhi Zhang, Xinyu Li, Joseph Tighe, S{\"o}ren
  Schwertfeger, Cyrill Stachniss, and Mu~Li.
\newblock Video contrastive learning with global context.
\newblock In \emph{Proceedings of the IEEE/CVF International Conference on
  Computer Vision}, pp.\  3195--3204, 2021.

\bibitem[Lester et~al.(2021)Lester, Al-Rfou, and Constant]{prompttuning}
Brian Lester, Rami Al-Rfou, and Noah Constant.
\newblock The power of scale for parameter-efficient prompt tuning.
\newblock \emph{arXiv preprint arXiv:2104.08691}, 2021.

\bibitem[Li et~al.(2021)Li, Wang, Peng, Song, Liu, Li, and
  Qiao]{li2021uniformer}
Kunchang Li, Yali Wang, Gao Peng, Guanglu Song, Yu~Liu, Hongsheng Li, and
  Yu~Qiao.
\newblock Uniformer: Unified transformer for efficient spatial-temporal
  representation learning.
\newblock In \emph{International Conference on Learning Representations}, 2021.

\bibitem[Li \& Liang(2021)Li and Liang]{li-liang-2021-prefix}
Xiang~Lisa Li and Percy Liang.
\newblock Prefix-tuning: Optimizing continuous prompts for generation.
\newblock In \emph{Proceedings of the 59th Annual Meeting of the Association
  for Computational Linguistics and the 11th International Joint Conference on
  Natural Language Processing (Volume 1: Long Papers)}, pp.\  4582--4597,
  Online, August 2021. Association for Computational Linguistics.
\newblock \doi{10.18653/v1/2021.acl-long.353}.
\newblock URL \url{https://aclanthology.org/2021.acl-long.353}.

\bibitem[Li et~al.(2022)Li, Wu, Fan, Mangalam, Xiong, Malik, and
  Feichtenhofer]{mvitv2}
Yanghao Li, Chao-Yuan Wu, Haoqi Fan, Karttikeya Mangalam, Bo~Xiong, Jitendra
  Malik, and Christoph Feichtenhofer.
\newblock Mvitv2: Improved multiscale vision transformers for classification
  and detection.
\newblock In \emph{2022 IEEE/CVF Conference on Computer Vision and Pattern
  Recognition (CVPR)}, pp.\  4794--4804, 2022.
\newblock \doi{10.1109/CVPR52688.2022.00476}.

\bibitem[Li et~al.(2018)Li, Li, and Vasconcelos]{diving48}
Yingwei Li, Yi~Li, and Nuno Vasconcelos.
\newblock Resound: Towards action recognition without representation bias.
\newblock In \emph{Proceedings of the European Conference on Computer Vision
  (ECCV)}, pp.\  513--528, 2018.

\bibitem[Lin et~al.(2019)Lin, Gan, and Han]{lin2019tsm}
Ji~Lin, Chuang Gan, and Song Han.
\newblock Tsm: Temporal shift module for efficient video understanding.
\newblock In \emph{Proceedings of the IEEE/CVF International Conference on
  Computer Vision}, pp.\  7083--7093, 2019.

\bibitem[Lin et~al.(2022)Lin, Geng, Zhang, Gao, de~Melo, Wang, Dai, Qiao, and
  Li]{frozenclip}
Ziyi Lin, Shijie Geng, Renrui Zhang, Peng Gao, Gerard de~Melo, Xiaogang Wang,
  Jifeng Dai, Yu~Qiao, and Hongsheng Li.
\newblock Frozen clip models are efficient video learners.
\newblock \emph{arXiv preprint arXiv:2208.03550}, 2022.

\bibitem[Liu et~al.(2021)Liu, Lin, Cao, Hu, Wei, Zhang, Lin, and
  Guo]{liu2021swin}
Ze~Liu, Yutong Lin, Yue Cao, Han Hu, Yixuan Wei, Zheng Zhang, Stephen Lin, and
  Baining Guo.
\newblock Swin transformer: Hierarchical vision transformer using shifted
  windows.
\newblock In \emph{Proceedings of the IEEE/CVF International Conference on
  Computer Vision}, pp.\  10012--10022, 2021.

\bibitem[Liu et~al.(2022)Liu, Ning, Cao, Wei, Zhang, Lin, and
  Hu]{liu2022videoswin}
Ze~Liu, Jia Ning, Yue Cao, Yixuan Wei, Zheng Zhang, Stephen Lin, and Han Hu.
\newblock Video swin transformer.
\newblock In \emph{Proceedings of the IEEE/CVF Conference on Computer Vision
  and Pattern Recognition}, pp.\  3202--3211, 2022.

\bibitem[Ni et~al.(2022)Ni, Peng, Chen, Zhang, Meng, Fu, Xiang, and
  Ling]{xclip}
Bolin Ni, Houwen Peng, Minghao Chen, Songyang Zhang, Gaofeng Meng, Jianlong Fu,
  Shiming Xiang, and Haibin Ling.
\newblock Expanding language-image pretrained models for general video
  recognition.
\newblock \emph{arXiv preprint arXiv:2208.02816}, 2022.

\bibitem[Qian et~al.(2021)Qian, Shao, Zhu, Li, and Jia]{qian2021blending}
Shengju Qian, Hao Shao, Yi~Zhu, Mu~Li, and Jiaya Jia.
\newblock Blending anti-aliasing into vision transformer.
\newblock \emph{Advances in Neural Information Processing Systems},
  34:\penalty0 5416--5429, 2021.

\bibitem[Qian et~al.(2022)Qian, Zhu, Li, Li, and Jia]{qian2022makes}
Shengju Qian, Yi~Zhu, Wenbo Li, Mu~Li, and Jiaya Jia.
\newblock What makes for good tokenizers in vision transformer?
\newblock \emph{IEEE Transactions on Pattern Analysis and Machine
  Intelligence}, 2022.

\bibitem[Qing et~al.(2022)Qing, Zhang, Huang, Wang, Wang, Lv, Gao, and
  Sang]{qing2022mar}
Zhiwu Qing, Shiwei Zhang, Ziyuan Huang, Xiang Wang, Yuehuan Wang, Yiliang Lv,
  Changxin Gao, and Nong Sang.
\newblock Mar: Masked autoencoders for efficient action recognition.
\newblock \emph{arXiv preprint arXiv:2207.11660}, 2022.

\bibitem[Radford et~al.(2021)Radford, Kim, Hallacy, Ramesh, Goh, Agarwal,
  Sastry, Askell, Mishkin, Clark, et~al.]{clip}
Alec Radford, Jong~Wook Kim, Chris Hallacy, Aditya Ramesh, Gabriel Goh,
  Sandhini Agarwal, Girish Sastry, Amanda Askell, Pamela Mishkin, Jack Clark,
  et~al.
\newblock Learning transferable visual models from natural language
  supervision.
\newblock In \emph{International Conference on Machine Learning}, pp.\
  8748--8763. PMLR, 2021.

\bibitem[Ryoo et~al.(2021)Ryoo, Piergiovanni, Arnab, Dehghani, and
  Angelova]{ryoo2021tokenlearner}
Michael~S Ryoo, AJ~Piergiovanni, Anurag Arnab, Mostafa Dehghani, and Anelia
  Angelova.
\newblock Tokenlearner: Adaptive space-time tokenization for videos.
\newblock In A.~Beygelzimer, Y.~Dauphin, P.~Liang, and J.~Wortman Vaughan
  (eds.), \emph{Advances in Neural Information Processing Systems}, 2021.
\newblock URL \url{https://openreview.net/forum?id=z-l1kpDXs88}.

\bibitem[Sevilla-Lara et~al.(2021)Sevilla-Lara, Zha, Yan, Goswami, Feiszli, and
  Torresani]{sevilla2021only}
Laura Sevilla-Lara, Shengxin Zha, Zhicheng Yan, Vedanuj Goswami, Matt Feiszli,
  and Lorenzo Torresani.
\newblock Only time can tell: Discovering temporal data for temporal modeling.
\newblock In \emph{Proceedings of the IEEE/CVF Winter Conference on
  Applications of Computer Vision}, pp.\  535--544, 2021.

\bibitem[Sun et~al.(2017)Sun, Shrivastava, Singh, and Gupta]{jft300m}
Chen Sun, Abhinav Shrivastava, Saurabh Singh, and Abhinav Gupta.
\newblock Revisiting unreasonable effectiveness of data in deep learning era.
\newblock In \emph{Proceedings of the IEEE international conference on computer
  vision}, pp.\  843--852, 2017.

\bibitem[Sung et~al.(2021)Sung, Nair, and Raffel]{fixsparsemask}
Yi-Lin Sung, Varun Nair, and Colin~A Raffel.
\newblock Training neural networks with fixed sparse masks.
\newblock \emph{Advances in Neural Information Processing Systems},
  34:\penalty0 24193--24205, 2021.

\bibitem[Tan et~al.(2021)Tan, Lei, Wolf, and Bansal]{tan2021vimpac}
Hao Tan, Jie Lei, Thomas Wolf, and Mohit Bansal.
\newblock Vimpac: Video pre-training via masked token prediction and
  contrastive learning.
\newblock \emph{arXiv preprint arXiv:2106.11250}, 2021.

\bibitem[Tong et~al.(2022)Tong, Song, Wang, and Wang]{tong2022videomae}
Zhan Tong, Yibing Song, Jue Wang, and Limin Wang.
\newblock Videomae: Masked autoencoders are data-efficient learners for
  self-supervised video pre-training.
\newblock \emph{arXiv preprint arXiv:2203.12602}, 2022.

\bibitem[Tran et~al.(2018)Tran, Wang, Torresani, Ray, LeCun, and Paluri]{r21d}
Du~Tran, Heng Wang, Lorenzo Torresani, Jamie Ray, Yann LeCun, and Manohar
  Paluri.
\newblock A closer look at spatiotemporal convolutions for action recognition.
\newblock In \emph{Proceedings of the IEEE Conference on Computer Vision and
  Pattern Recognition}, pp.\  6450--6459, 2018.

\bibitem[Wang et~al.(2021{\natexlab{a}})Wang, Xing, and
  Liu]{wang2021actionclip}
Mengmeng Wang, Jiazheng Xing, and Yong Liu.
\newblock Actionclip: A new paradigm for video action recognition.
\newblock \emph{arXiv preprint arXiv:2109.08472}, 2021{\natexlab{a}}.

\bibitem[Wang et~al.(2022{\natexlab{a}})Wang, Chen, Wu, Chen, Dai, Liu, Jiang,
  Zhou, and Yuan]{wang2022bevt}
Rui Wang, Dongdong Chen, Zuxuan Wu, Yinpeng Chen, Xiyang Dai, Mengchen Liu,
  Yu-Gang Jiang, Luowei Zhou, and Lu~Yuan.
\newblock Bevt: Bert pretraining of video transformers.
\newblock In \emph{Proceedings of the IEEE/CVF Conference on Computer Vision
  and Pattern Recognition}, pp.\  14733--14743, 2022{\natexlab{a}}.

\bibitem[Wang et~al.(2021{\natexlab{b}})Wang, Xie, Li, Fan, Song, Liang, Lu,
  Luo, and Shao]{wang2021pyramidvit}
Wenhai Wang, Enze Xie, Xiang Li, Deng-Ping Fan, Kaitao Song, Ding Liang, Tong
  Lu, Ping Luo, and Ling Shao.
\newblock Pyramid vision transformer: A versatile backbone for dense prediction
  without convolutions.
\newblock In \emph{Proceedings of the IEEE/CVF International Conference on
  Computer Vision}, pp.\  568--578, 2021{\natexlab{b}}.

\bibitem[Wang et~al.(2022{\natexlab{b}})]{beitv3}
Wenhui Wang et~al.
\newblock {Image as a Foreign Language: BEiT Pretraining for All Vision and
  Vision-Language Tasks}.
\newblock \emph{arXiv preprint arXiv:2208.10442}, 2022{\natexlab{b}}.

\bibitem[Wei et~al.(2022)Wei, Fan, Xie, Wu, Yuille, and
  Feichtenhofer]{maskfeat}
Chen Wei, Haoqi Fan, Saining Xie, Chao-Yuan Wu, Alan Yuille, and Christoph
  Feichtenhofer.
\newblock Masked feature prediction for self-supervised visual pre-training.
\newblock In \emph{Proceedings of the IEEE/CVF Conference on Computer Vision
  and Pattern Recognition}, pp.\  14668--14678, 2022.

\bibitem[Xie et~al.(2022)Xie, Zhang, Cao, Lin, Bao, Yao, Dai, and
  Hu]{xie2022simmim}
Zhenda Xie, Zheng Zhang, Yue Cao, Yutong Lin, Jianmin Bao, Zhuliang Yao,
  Qi~Dai, and Han Hu.
\newblock Simmim: A simple framework for masked image modeling.
\newblock In \emph{Proceedings of the IEEE/CVF Conference on Computer Vision
  and Pattern Recognition}, pp.\  9653--9663, 2022.

\bibitem[Yan et~al.(2022)Yan, Xiong, Arnab, Lu, Zhang, Sun, and
  Schmid]{yan2022multiview}
Shen Yan, Xuehan Xiong, Anurag Arnab, Zhichao Lu, Mi~Zhang, Chen Sun, and
  Cordelia Schmid.
\newblock Multiview transformers for video recognition.
\newblock In \emph{Proceedings of the IEEE/CVF Conference on Computer Vision
  and Pattern Recognition}, pp.\  3333--3343, 2022.

\bibitem[Yang et~al.(2021)Yang, Zhu, Mendieta, Wang, Balakrishnan, Lee, Han,
  Shah, and Chen]{yang2021mutualnet}
Taojiannan Yang, Sijie Zhu, Matias Mendieta, Pu~Wang, Ravikumar Balakrishnan,
  Minwoo Lee, Tao Han, Mubarak Shah, and Chen Chen.
\newblock Mutualnet: Adaptive convnet via mutual learning from different model
  configurations.
\newblock \emph{IEEE Transactions on Pattern Analysis and Machine
  Intelligence}, 45\penalty0 (1):\penalty0 811--827, 2021.

\bibitem[Yu et~al.(2021)Yu, Naik, Backurs, Gopi, Inan, Kamath, Kulkarni, Lee,
  Manoel, Wutschitz, et~al.]{fl2}
Da~Yu, Saurabh Naik, Arturs Backurs, Sivakanth Gopi, Huseyin~A Inan, Gautam
  Kamath, Janardhan Kulkarni, Yin~Tat Lee, Andre Manoel, Lukas Wutschitz,
  et~al.
\newblock Differentially private fine-tuning of language models.
\newblock In \emph{International Conference on Learning Representations}, 2021.

\bibitem[Yuan et~al.(2021{\natexlab{a}})Yuan, Chen, Wang, Yu, Shi, Jiang, Tay,
  Feng, and Yan]{yuan2021tokenstotoken}
Li~Yuan, Yunpeng Chen, Tao Wang, Weihao Yu, Yujun Shi, Zi-Hang Jiang,
  Francis~EH Tay, Jiashi Feng, and Shuicheng Yan.
\newblock Tokens-to-token vit: Training vision transformers from scratch on
  imagenet.
\newblock In \emph{Proceedings of the IEEE/CVF International Conference on
  Computer Vision}, pp.\  558--567, 2021{\natexlab{a}}.

\bibitem[Yuan et~al.(2021{\natexlab{b}})Yuan, Chen, Chen, Codella, Dai, Gao,
  Hu, Huang, Li, Li, et~al.]{yuan2021florence}
Lu~Yuan, Dongdong Chen, Yi-Ling Chen, Noel Codella, Xiyang Dai, Jianfeng Gao,
  Houdong Hu, Xuedong Huang, Boxin Li, Chunyuan Li, et~al.
\newblock Florence: A new foundation model for computer vision.
\newblock \emph{arXiv preprint arXiv:2111.11432}, 2021{\natexlab{b}}.

\bibitem[Zhai et~al.(2022)Zhai, Kolesnikov, Houlsby, and
  Beyer]{zhai_cvpr2022_scalingvit}
Xiaohua Zhai, Alexander Kolesnikov, Neil Houlsby, and Lucas Beyer.
\newblock {Scaling Vision Transformers}.
\newblock In \emph{Proceedings of the IEEE/CVF Conference on Computer Vision
  and Pattern Recognition}, 2022.

\bibitem[Zhang et~al.(2021{\natexlab{a}})Zhang, Yu, Fifty, Han, Dai, Pang, and
  Sha]{cover}
Bowen Zhang, Jiahui Yu, Christopher Fifty, Wei Han, Andrew~M Dai, Ruoming Pang,
  and Fei Sha.
\newblock Co-training transformer with videos and images improves action
  recognition.
\newblock \emph{arXiv preprint arXiv:2112.07175}, 2021{\natexlab{a}}.

\bibitem[Zhang et~al.(2021{\natexlab{b}})Zhang, Li, Liu, Shuai, Zhu, Brattoli,
  Chen, Marsic, and Tighe]{zhang2021vidtr}
Yanyi Zhang, Xinyu Li, Chunhui Liu, Bing Shuai, Yi~Zhu, Biagio Brattoli, Hao
  Chen, Ivan Marsic, and Joseph Tighe.
\newblock Vidtr: Video transformer without convolutions.
\newblock In \emph{Proceedings of the IEEE/CVF International Conference on
  Computer Vision}, pp.\  13577--13587, 2021{\natexlab{b}}.

\bibitem[Zhao et~al.(2022)Zhao, Du, Li, Li, and Liu]{fl1}
Haodong Zhao, Wei Du, Fangqi Li, Peixuan Li, and Gongshen Liu.
\newblock Reduce communication costs and preserve privacy: Prompt tuning method
  in federated learning.
\newblock \emph{arXiv preprint arXiv:2208.12268}, 2022.

\bibitem[Zhong et~al.(2020)Zhong, Zheng, Kang, Li, and Yang]{randomerasing}
Zhun Zhong, Liang Zheng, Guoliang Kang, Shaozi Li, and Yi~Yang.
\newblock Random erasing data augmentation.
\newblock In \emph{Proceedings of the AAAI conference on artificial
  intelligence}, volume~34, pp.\  13001--13008, 2020.

\bibitem[Zhou et~al.(2021)Zhou, Wei, Wang, Shen, Xie, Yuille, and
  Kong]{zhou2021ibot}
Jinghao Zhou, Chen Wei, Huiyu Wang, Wei Shen, Cihang Xie, Alan Yuille, and Tao
  Kong.
\newblock ibot: Image bert pre-training with online tokenizer.
\newblock \emph{arXiv preprint arXiv:2111.07832}, 2021.

\bibitem[Zhu et~al.(2019)Zhu, Lan, Newsam, and Hauptmann]{zhu2019hidden}
Yi~Zhu, Zhenzhong Lan, Shawn Newsam, and Alexander Hauptmann.
\newblock Hidden two-stream convolutional networks for action recognition.
\newblock In \emph{Computer Vision--ACCV 2018: 14th Asian Conference on
  Computer Vision, Perth, Australia, December 2--6, 2018, Revised Selected
  Papers, Part III 14}, pp.\  363--378. Springer, 2019.

\bibitem[Zhu et~al.(2020)Zhu, Li, Liu, Zolfaghari, Xiong, Wu, Zhang, Tighe,
  Manmatha, and Li]{zhu2020videosurvey}
Yi~Zhu, Xinyu Li, Chunhui Liu, Mohammadreza Zolfaghari, Yuanjun Xiong, Chongruo
  Wu, Zhi Zhang, Joseph Tighe, R.~Manmatha, and Mu~Li.
\newblock A comprehensive study of deep video action recognition.
\newblock \emph{arXiv preprint arXiv:2012.06567}, 2020.

\bibitem[Zolfaghari et~al.(2021)Zolfaghari, Zhu, Gehler, and
  Brox]{zolfaghari2021crossclr}
Mohammadreza Zolfaghari, Yi~Zhu, Peter Gehler, and Thomas Brox.
\newblock Crossclr: Cross-modal contrastive learning for multi-modal video
  representations.
\newblock In \emph{Proceedings of the IEEE/CVF International Conference on
  Computer Vision}, pp.\  1450--1459, 2021.

\end{thebibliography}
	\bibliographystyle{iclr2023_conference}
	
	\appendix
	
	
	\section{Implementation details}
	\label{app:implementation}
	\subsection{Kinetics 400/700}
	We add spatial/temporal/joint adaptation in every ViT block as shown in Fig.\ \ref{fig:adapter}. The bottleneck ratios of all adapters are set to 0.25 and the scaling factor is set to 0.5. The first FC layer in Adapters is randomly initialized and the second FC layer is initialized to zero. In this way, the adapted model is close to the pre-trained model at the beginning of training. We largely follow the training settings and data augmentations in \citet{liu2022videoswin}. Specifically, the model is trained for 30 epochs using AdamW \citep{kingma2014adam} optimizer with a batchsize of 64. The base learning rate is 3e-4 and weight decay is 5e-2. The learning rate is warmed up from 0 in the first 3 epochs and then decays following a cosine schedule. The stochastic depth rate is 0.2 for both ViT-B and ViT-L. For inference, we sample three clips along the temporal dimension. The final performance is evaluated by the ensemble of three views. We evaluate the model on 8, 16, 32 frames and the sampling interval is 16, 8, 4, respectively.
	
	\subsection{Something-something-v2}
	We add spatial/temporal/joint adaptation in every ViT block as shown in Fig.\ \ref{fig:adapter}. We additionally add one adapter before T-MSA to enhance the temporal modeling. The bottleneck ratios of all adapters are set to 0.25 and the scaling factor is set to 0.5. We follow \citet{liu2022videoswin} to use stronger data augmentations including label smoothing, RandAugment \citep{cubuk2020randaugment} and random erasing \citep{randomerasing}. The model is trained for 50 epochs using AdamW \citep{kingma2014adam} optimizer. The other training settings are the same as Kinetics-400. We uniformly sample 8, 16, 32 frames in the experiments. For inference, we sample three spatial crops. The final performance is evaluated by the ensemble of three views.
	
	\subsection{Diving-48}
	We add spatial/temporal/joint adaptation in every ViT block as shown in Fig.\ \ref{fig:adapter}. The bottleneck ratios of all adapters are set to 0.25 and the scaling factor is set to 0.5. The model is trained for 50 epochs. The other training settings and data augmentations are the same as K400. We uniformly sample 8, 16, 32 frames in the experiments. For inference, we only sample 1 temporal clip.
	
	\section{Visualization}
	\begin{figure}[t]
		\centering
		\includegraphics[width=0.99\linewidth]{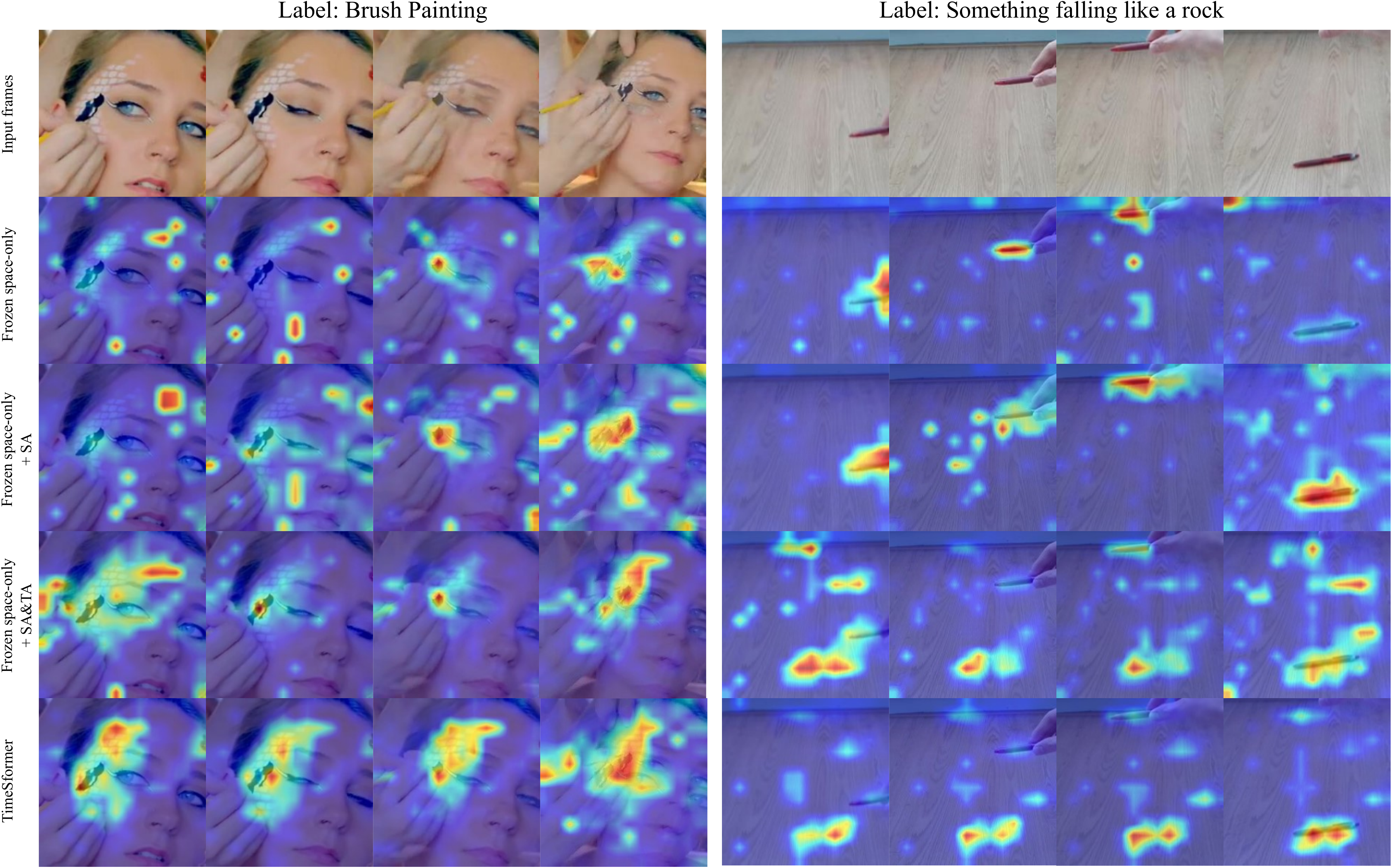}
		\caption{Attention map visualizations of  AIM variants and the full finetuned TimeSformer. With the help of temporal adaptation (TA), our method is able to focus on motion salient regions which helps to make a correct prediction.}
		\label{fig:visual}
	\end{figure}
	
	In this section, we present the attention map visualizations of the frozen space-only model, Spatial Adaptation (SA) model, Spatial Adaptation plus Temporal Adaptation (TA) model, and the full finetuned TimeSformer.
	
	On Fig. \ref{fig:visual} left, we visualize an action ``Brush Painting'' from Kinetics-400 dataset. 
	We can see that the attention maps of the frozen space-only model are very scattered, and it doesn't attend to the brush region in the first two frames. 
	Adding SA enhances the attention on the brush, but the model still focuses on areas that are unrelated to the action. 
	Further adding TA helps the model to learn temporal information. 
	We can see that the model now focuses more on the brush painting area, which is similar to what full finetuned TimeSformer does. 
	
	On Fig. \ref{fig:visual} right, we visualize an action ``Something falling like a rock'' from Something-Something-v2 dataset. 
	To correctly recognize this action, the model needs to learn how the object moves in the input frames. 
	We first observe that both the frozen space-only model and SA model have good attention on the object, but they fail to model the movement of the object which leads to wrong prediction. 
	In contrast, TA helps the model to learn the relationship among input frames.
	The attention map shows that the model not only focuses on the object but also learns the track of the object. 
	Instead, TimeSformer always attends to the bottom region without showing the object path. 
	
	\section{Per-class analysis}
	\begin{figure}
		\centering
		\includegraphics[width=0.75\linewidth]{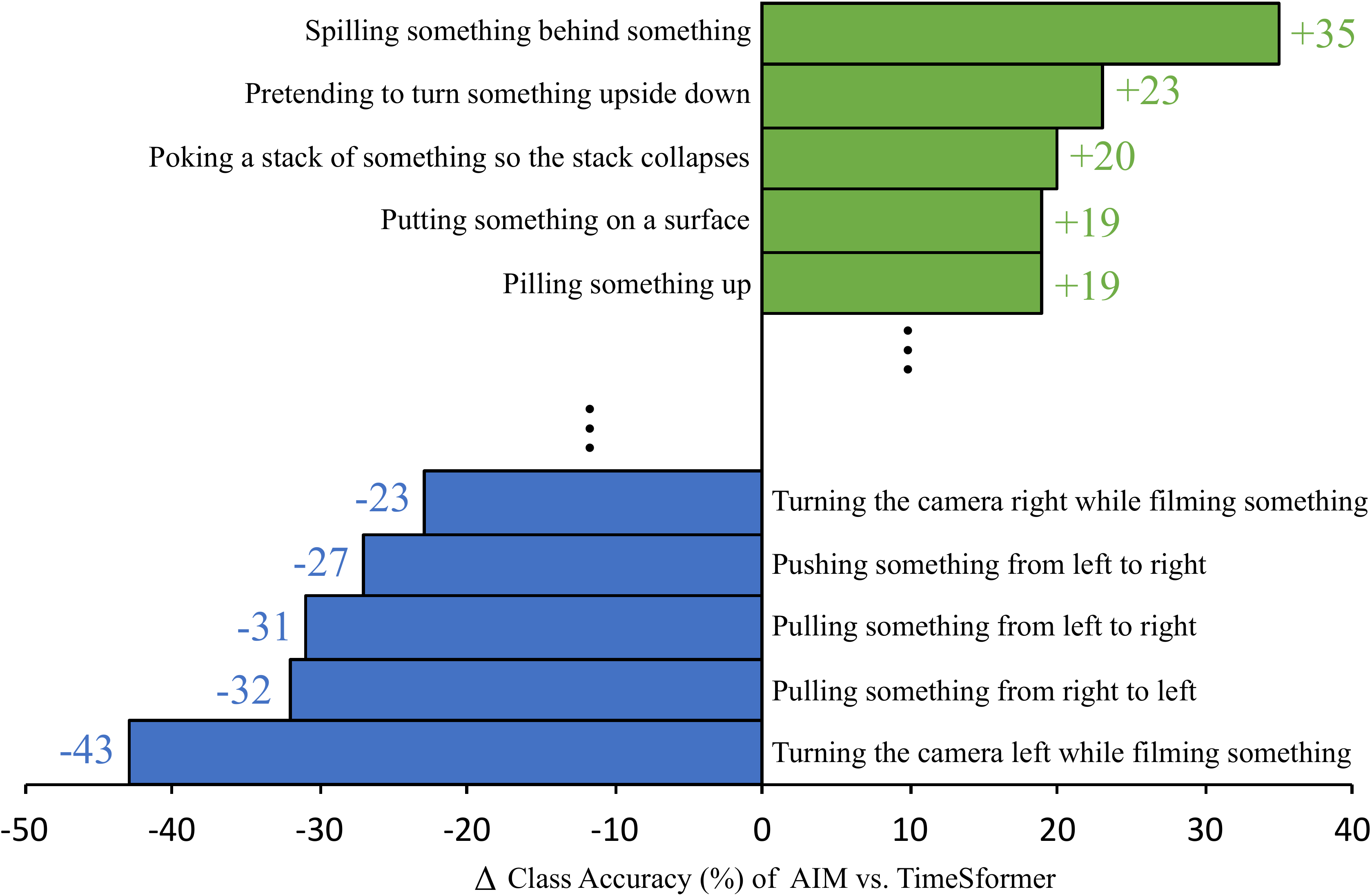}
		\caption{The figure shows the differences of each class's accuracy of AIM and TimeSformer on Something-Something-v2. Here we only plot the top-5 and bottom-5 classes.}
		\label{fig:classaccuracy}
	\end{figure}
	
	In Tab. \ref{tab:ssv2}, we show that AIM still falls behind some SoTA full finetuned video models on the ``temporal-heavy'' Something-Something-v2 (SSv2) dataset. We conjecture one reason is that simply reusing the image pre-trained self-attention for temporal modeling may not be able to fully capture the complicated temporal information in some nuanced action classes in SSv2. 
	To provide further insights, we compute the per-class accuracy differences of AIM and TimeSformer on SSv2 and show the top-5 and bottom-5 classes in Fig. \ref{fig:classaccuracy}. 
	We can see that the classes where AIM performs better are normal action classes with decent motion. 
	The classes where AIM performs worse are those with minor differences (\eg, ``Pulling something from left to right'' vs. ``Pulling something from right to left''). 
	In order to tell these actions apart, the model needs to distinguish between the nuances, especially in motion. 
	Given most of model parameters are frozen in our method, AIM may lack the capacity to capture such complex temporal information.
	
	\section{More comparisons of training cost}
	In Table \ref{tab:pretrain}, we demonstrate the training efficiency of AIM based on ViT-B and Swin-B backbones. In this section, we show more comparisons with full finetuned baselines based on ViT-L and Swin-L backbones. The results are shown in Table \ref{tab:memorycost}. We can see that TimeSformer with a ViT-L backbone needs 21.2G GPU memory, and VideSwin with a Swin-L backbone cannot fit into an 8 Tesla V100 32G GPU server. In both cases, AIM can significantly reduce the memory usage to 14.3G and 13.7G, respectively. This makes large model training more memory-friendly (runnable on most GPUs with 15G memory and more) , and thus more affordable for most researchers and practioners.
	
	Furthermore, beyond memory saving, optimizing number of tunable parameters has potential benefits in other applications such as communication-efficient distributed learning (e.g., federated learning where the tunable model parameters are communicated between the central server and local clients) and privacy preserving federated learning \cite{fl1, fl2}. Tuning less parameters could also be beneficial when the downstream data is limited because fully finetuning a large model on limited data may suffer from serious overfitting.  This can be observed from the results in Fig. \ref{fig:dataefficiency}  where AIM obtains larger accuracy improvements over the full finetuned baseline when there is only small amount of training data.

	\begin{table}[t]
		\caption{Comparisons of the training memory cost of AIM and full finetuned models based on large image pre-trained backbones. AIM significantly reduces the memory cost and makes large model training easier.}
		\begin{center}
			\begin{tabular}{l|cc}
				\hline
				Model & Backbone &  Mem (G)   \\
				\hline
				TimeSformer \cite{timesformer} & ViT-L & 21.2 \\
				AIM & ViT-L & 14.3 \\
				\hline
				VideoSwin \cite{liu2022videoswin} & Swin-L & Out of Memory \\
				AIM & Swin-L & 13.7 \\
				\hline
			\end{tabular}
		\end{center}
		\label{tab:memorycost}
	\end{table}

	\section{Comparison to EVL under different pre-trained datasets}
	In this section, we compare AIM with EVL \cite{frozenclip}, which is the most recent SoTA image-to-video efficient finetuning method based on frozen pre-trained ViT. As shown in the Table \ref{tab:compareevl}, AIM consistently outperforms EVL under both IN-21K and CLIP pre-training as well. And AIM uses considerably less tunable number of parameters than EVL.

	\begin{table}[t]
		\caption{Comparisons with EVL under different pre-trained datasets. AIM outperforms EVL under different pre-training and uses less number of tunable parameters.}
		\begin{center}
			\begin{tabular}{l|cc|cccc}
				\hline
				Model & Backbone & Pretrain & \makecell{Tunable \\ Param (M)} &  Mem (G) & Time (H) & Top-1  \\
				\hline
				EVL \cite{frozenclip} & ViT-B & IN-21K & 36.3 & 4.2 & 29 & 75.4 \\
				AIM & ViT-B & IN-21K & 11 & 7 & 15 & \textbf{78.8} \\
				\hline
				EVL \cite{frozenclip} & ViT-B & CLIP & 36.3 & 4.2 & 29 & 82.9 \\
				AIM & ViT-B & CLIP & 11 & 7 & 15 & \textbf{83.9} \\
				\hline
			\end{tabular}
		\end{center}
		\label{tab:compareevl}
	\end{table}
	
	\section{Pseudo-code of the adapted ViT block}
	As explained in the paper, AIM is effective and simple to implement. In Algorithm \ref{alg:aim}, we show the PyTorch style pseudo-code on how to apply AIM to a ViT block.
	
	\begin{algorithm}[h!]
		\caption{Pseudo-code of an adapted ViT block}
		\label{alg:aim}
		\begin{minted}[fontsize=\footnotesize]{python}
			class TransformerBlock():
			
			def __init__(self, dim, num_head, mlp_ratio, scale):
			## Layers in the original ViT block
			self.attn = MultiheadAttention(dim, num_head)
			self.norm1 = LayerNorm(dim)
			self.mlp = MLP(dim, mlp_ratio)
			self.norm2 = LayerNorm(dim)
			
			## Adapters
			self.s_adapter = Adapter(dim)
			self.t_adapter = Adapter(dim)
			self.mlp_adapter = Adapter(dim)
			self.scale = scale
			
			def forward(x):
			## x in shape [N+1, BT, D]
			
			## temporal adaptation
			xt = rearrange(x, 'n (b t) d -> t (b n) d', t=num_frames)
			xt = self.t_adapter(self.attn(self.norm1(x)))
			xt = rearrange(x, 't (b n) d -> n (b t) d', n=num_patches)
			x = x + xt
			
			## spatial adaptation
			x = x + self.s_adapter(self.attn(self.norm1(x)))
			
			## joint adaptation
			x_norm = self.norm2(x)
			x = x + self.mlp(x_norm) + self.scale * self.mlp_adapter(x_norm)
			
			return x
			
		\end{minted}
	\end{algorithm}

\end{document}